\documentclass[11pt]{article}
\usepackage{coling2018}
\usepackage{times}
\usepackage{url}
\usepackage{latexsym}
\usepackage{amsmath}
\usepackage{acronym}
\usepackage{verbatim}
\usepackage{graphicx}
\usepackage{layout}
\usepackage{comment}
\usepackage{array}

\acrodef{BLEU}{Bilingual Evaluation Understudy}
\acrodef{DP}{Dynamic Programming}
\acrodef{EM}{Expectation Maximisation}
\acrodef{SMT}{Statistical Machine Translation}
\acrodef{MT}{Machine Translation}
\acrodef{ML} {Machine Learning}
\acrodef{RST}{Rhetorical Structure Theory}
\acrodef{EDU}{elementary discourse units}
\acrodef{WSD}{Word Sense Disambiguation}
\acrodef{HT}{Human Translation}
\acrodef{PE}{post-edited}
\acrodef{HMM}{Hidden Markov Model}
\acrodefindefinite{HMM}{an}{a}
\acrodef{LDA}{Latent Dirichlet Allocation}
\acrodef{LM}{Language Model}
\acrodef{TM}{Translation Model}
\acrodefindefinite{LM}{an}{a}
\acrodef{PDTB}{Penn Discourse Tree Bank}
\acrodef{MERT}{Minimum Error Rate Training}
\acrodefindefinite{MERT}{an}{a}
\acrodef{ML}{Machine Learning}
\acrodefindefinite{ML}{an}{a}
\acrodef{MLE}{Maximum Likelihood Estimation}
\acrodefindefinite{MLE}{an}{a}
\acrodef{MT}{Machine Translation}
\acrodefindefinite{MT}{an}{a}
\acrodef{POS}{Part-of-Speech}
\acrodef{QE}{Quality Estimation}
\acrodef{REF}{Reference}
\acrodef{RST}{Rhetorical Structure Theory}
\acrodef{ST}{Source Text}
\acrodef{SL}{Source Language}
\acrodefindefinite{SL}{an}{a}
\acrodef{SMT}{Statistical Machine Translation}
\acrodefindefinite{SMT}{an}{a}
\acrodef{LSTM}{Long-Short Term Memory}
\acrodef{TT}{Target Text}
\acrodef{TL}{Target Language}	
\acrodef{NLG}{Natural Language Generation}
\acrodef{HRNNLM}{Hierarchical Recurrent Neural Network Language Model}
\acrodef{RNNLM}{Recurrent Neural Network Language Model}
\acrodef{HTER}{Human-targeted Translation Error Rate}	
\acrodef{MAE}{mean absolute error}
\acrodef{PBMT}{Phrase Based Machine Translation}
\acrodef{NMT}{Neural Machine Translation}
\acrodef{MAE}{mean absolute error}
\acrodef{UD}{Universal Dependencies}

\title{Assessing Crosslingual Discourse Relations in Machine Translation}

\author{Karin Sim Smith\qquad\qquad Lucia Specia\\
	Department of Computer Science \\
	The University of Sheffield \\
	Sheffield, UK \\
	{\tt karinsim@gmail.com, l.specia@sheffield.ac.uk}\\}

\date{}

\begin{document}
	\maketitle
	\begin{abstract}
		In an attempt to improve overall translation quality, there has been an increasing focus on integrating more linguistic elements into \ac{MT}.
		While significant progress has been achieved, especially recently with neural models, automatically evaluating the output of such systems is still an open problem. 
		Current practice in \ac{MT} evaluation relies on a single reference translation, even though there are many ways of translating a particular text, and it tends to disregard higher level information such as discourse. 
		We propose a novel approach that assesses the translated output based on the \emph{source} text rather than the reference translation, and measures the extent to which the semantics of the discourse elements (discourse relations, in particular) in the source text are preserved in the \ac{MT} output.
		The challenge is to detect the discourse relations in the \emph{source} text and determine whether these relations are correctly transferred crosslingually to the target language -- \emph{without} a reference translation.
		This methodology could be used independently for discourse-level evaluation, or as a component in other metrics, at a time where substantial amounts of \ac{MT} are online and would benefit from evaluation where the \emph{source} text serves as a benchmark.                                                                                                            
	\end{abstract}

	\section{Introduction}
	Despite the fact that discourse relations have long been recognised as crucial to the proper understanding of a text 
	\cite{grimes,longacre}, current \ac{MT} systems often fail to properly handle discourse relations for various reasons, such as incorrect word alignments, the presence of multi word expressions as discourse markers, and the prevalence of ambiguous or implicit discourse markers. \ac{MT} systems generally do not take account of discourse relations explicitly, and discourse connectives are simply treated as any other words to be translated.
	
	Previous research on assessing discourse relations in \ac{MT} has covered work incorporating discourse structure in the evaluation of \ac{MT} output by comparing the discourse tree structure of the \ac{MT} to that of the gold standard (reference translation) \cite{Guzman}, or on a (also reference-based) discourse-connective specific metric \cite{Hajlaoui:2013}. 		
	While using a reference for evaluating a candidate translation is the norm, and has the benefit of allowing automatic evaluation, it is an inflexible and potentially restrictive way of evaluating translations. 
	There can be a number of ways a text can be correctly translated, but usually only one reference is available, and evaluation is made by shallow comparisons with that one reference translation.
		This is a problem that is evident in practice, when mistranslations can lead to business losses \cite{Levin}. 
		
	Moreover the prerequisite of a gold standard reference for evaluation is limiting, as evaluations can only be performed on small, pre-defined test sets. For the large amount of online \ac{MT} that takes place nowadays there is no reference, and there is therefore no way of verifying the correctness of the output.
	This poses a real problem, as illustrated recently
	when Facebook had to issue an apology over a mistranslation which had led to someone's arrest\footnote{		https://www.theguardian.com/technology/2017/oct/24/facebook-palestine-israel-translates-good-morning-attack-them-arrest}.
	The field of \ac{NLG} has a similar problem, and researchers are reaching a similar conclusion on the need for reference-less evaluation based on \emph{source meaning} \cite{novikova_HW,Dusek}. 
We believe that one way to solve this problem is to measure with respect to the \emph{source} text, as a human translator would. 
	
	By way of experiment, we attempt to establish how a particular discourse relation in the source text should be rendered in the target text, 
	and then to evaluate how well this is translated in the \ac{MT} output. This is not in comparison with a gold standard, but based on whether the \ac{MT} output carries the intended meaning of the source text. 
	In that respect, our task is more challenging than the aforementioned metric: instead of performing string matching to detect the presence of certain connectives in the \ac{MT} output and reference translation, we need to detect the discourse relations in the \emph{source} text and determine whether these relations are correctly transferred crosslingually to the target language -- \emph{without} a reference translation.
	
	We assume that the semantics of a discourse relation should transfer from source to target language, as this has been broadly established \cite{daCunha,Laali2014}. In other words, while the actual segmentation into discourse units may vary from language to language \cite{Mitkov}, the meaning of the actual discourse relation is constant across languages. Therefore, on the basis that discourse relations are semantically similar across languages, we assess how well \emph{explicit} discourse relations are transferred from the source to the target language. This will work for the languages we have chosen: French, and English, where we take French as the \ac{ST} and English as the \ac{TT}. In these languages, discourse relations often take the form of lexical cues, or discourse connectives, which signal the existence of a particular discourse relation. This is the case particularly for the lexically-based \ac{PDTB} \cite{Prasad:2008}, and to a lesser extent the hierarchical \ac{RST} \cite{Knott93}, as illustrated by work on lexical cues in \ac{RST} \cite{Khazaei}. 
	
	Compiling static lists of equivalent discourse markers in two languages is a cumbersome approach: given the variety of discourse markers in both languages, the range of relationship each can encode, and the fact that there are many alternative lexicalisations for discourse connectives \cite{Prasad:2010}. 
	Moreover, as established in previous research on  crosslingual discourse relations \cite{Meyer,MeyerPopescu,Li:2014b}, there can be mismatches due to ambiguous discourse markers. We therefore train crosslingual discourse embeddings which we hope will capture equivalences in discourse connectives across languages in a more flexible manner due to the context which they encapsulate. 
	Essentially a metric, this experiment incorporates a likelihood score from the crosslingual embeddings, in addition to a weighted score for the correctness of the particular discourse relation. 
	We evaluate the latter by assessing how well different outputs render the discourse relation as represented in the source text.
	We first compare the scores of \ac{MT} output over that of a post-edited version of the same, and find that our metric scores the \ac{PE} greater or equal to the \ac{MT} for 78\% of the documents. It is important to note that we are only focussing on one aspect of the output, that of the discourse relation, and in many cases that will have been adequately rendered by the \ac{MT}, resulting in a draw. Moreover, the \ac{MT} output tends to achieve a higher likelihood score from the embeddings, which are probabilistic by nature. 
	We also evaluate system submissions from the WMT14 campaign, finding that comparing our rankings to the human rankings results in some notable exceptions regarding rules-based systems.
	Our metric is novel in that it evaluates the \ac{MT} directly against the source text. In addition, it uses crosslingual multiword embeddings trained on discourse connectives and incorporates discourse relation mappings between source and target texts.

	In Section \ref{related} we describe related work, before detailing  the pipeline to build our metric (Section \ref{methodology}), including how we created the crosslingual embeddings to specifically track the discourse connectives, the methodology employed to evaluate the target relations based on the source text and our overall scoring metric. In Section \ref{DRresults} we present the results from our experiments.


	\section{Related Work}\label{related}

	The aforementioned work on crosslingual discourse relations \cite{Meyer,MeyerPopescu,Li:2014b} investigated disambiguating ambiguous connectives, using translation spotting and subsequent annotation of phrase tables to try and improve \ac{MT} output.	
	As mentioned, \newcite{Guzman} used discourse structures to evaluate \ac{MT} output:  
	they hypothesize that the discourse structure of good translations will have similar discourse relations. They parse both \ac{MT} output and reference translation for discourse relations and use tree kernels to compare \ac{HT} and \ac{MT} discourse tree structures. They show that scores derived from discourse structure are complementary to existing metrics. We display their scores for purposes of comparison, in Section \ref{DRresults}.
	
	Closest to our work is the ACT metric \cite{Hajlaoui:2013}, but it focuses on a narrow selection of ambiguous connectives, and uses a reference translation. 
	Taking seven ambiguous connectives, it scores the \ac{MT} output based on whether it has the same or equivalent connectives to the ones used in the reference. It uses a static list of equivalents for each of these seven connectives, and judges whether the sense of the target connective is compatible with that of the source based on how it has been translated in the \emph{reference}.
	Our approach is broader in that instead of a small list of static mappings of corresponding discourse connectives in both languages, we use crosslingual embeddings to identify correspondences between discourse connectives in two languages. 
	Our metric not only avoids the need for a reference translation, but is more realistic as there can be many ways of translating a text, all equally valid,  and in real world scenarios there is no reference to compare with.
	
	\section{Methodology}\label{methodology} 
	
\begin{figure}
		\includegraphics[width=\textwidth]{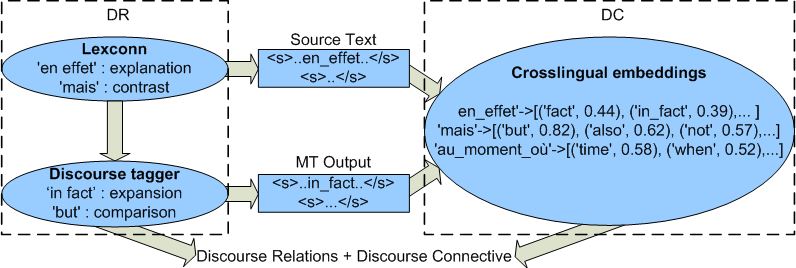}
		\caption[Dis-Score components]{Dis-Score incorporates a discourse relation component and a discourse connective component. The DR component (left) uses the LexConn definitions combined with syntax rules to determine the French connective, comparing to the relation derived by the tagger for the English. The DC component (right) incorporates the probability a particular French connective is translated as the one found in the \ac{MT} output. }\label{fig:diagram}
	\end{figure}

	As can be seen from the diagram in Figure \ref{fig:diagram}, our discourse score is composed of two components: Discourse Relation (DR) and Discourse Connective (DC).
	For measuring the correctness of the Discourse Connective, we take the likelihood score given from pretrained word embeddings for translation of the cue. 
	In Discourse Relation, the semantics of the discourse relations themselves are estimated by comparing (a) the discourse relation of the source text, as assessed from usages defined in LexConn \cite{roze_Lexconn}, a French lexicon of 328 discourse connectives, with (b) the relation of the target text, as assessed via a discourse tagger for English \cite{PitlerNenkova:2009}.
	We establish whether the source text discourse relation (identified via syntactic lexical cues combined with LexConn) corresponds to the discourse relation present in the MT output (based on the English discourse tagger). These two components together give an indication of how well the explicit discourse relation is transferred from source to target text. 
	We use the Stanford CoreNLP toolkit \cite{manning-EtAl:2014} to parse both the French (in order to ascertain the discourse usage of a potential cue) and the English \ac{MT} output (to provide as input to the English discourse tagger of \cite{PitlerNenkova:2009}). We detail these two components in Sections \ref{sec:connectives} and \ref{sec:relations}.
	This methodology could be applied to other language pairs, but by its very nature the implementation is language-specific. For language pairs where one or both languages have a substantial amount of implicit discourse relations, a discourse parser would be needed instead of cues.
	
	\subsection{Datasets}\label{sec:dr_datasets}
	We used French-English as our language pair, as the quality of the MT output needs to be of a certain level before attempting to discern transfer of discourse relations. 
	In addition, we needed a dataset with post-edited translations for evaluation purposes. Post-edited translations consist of the \ac{MT} output with the minimum amount of editing necessary in order to make it an acceptable translation. While the remit given to the posteditor varies, in general, the changes should be corrections involving basic semantic and grammatical errors, not stylistic changes.
	As we discuss next, 
	two types of data are needed: 
	a large amount of human-translated parallel data to build word embeddings models, and a lesser amount of \ac{MT} output parallel data for testing of the models. 
	\paragraph{Training Data}
	For training our discourse-specific bilingual embeddings we require a parallel corpus.
	In order to ensure that the correct use of discourse connectives is captured in the training of our embeddings, 
	we use a filtered version of the Europarl corpus \cite{Europarl}, as provided by IDIAP\footnote{https://www.idiap.ch/dataset/europarl-direct} \cite{Cartoni}. This consists of a filtered source text to ensure that only the excerpts which were originally written in French (in our case) are taken as source. 
	The parallel text is then formed by the original French sentences and their English reference translation. 
	Given that many discourse markers may be composed of several words (see Section \ref{sec:connectives} for a full explanation),  
	we hyphenate the discourse cues in the training data -- as per previous work on training phrases \cite{mikolov2013distributed}. 
	We also train the embeddings with a non-hyphenated version, and with the full Europarl dataset \cite{Europarl}. This results in three embedding models (hyphenated, non-hyphenated, and full Europarl) (Section \ref{sec:connectives}). 

	\paragraph{LIG Test Data}
	As our main test data we use the LIG corpus \cite{Potet:2012} of French-English translations. 
	This dataset was chosen since it includes a \ac{PE} version, suitable for evaluation in our task. 
	In all it comprises 361 parallel documents, a total of 10,755 sentence tuples: $<$FR, MT, PE, HT$>$, including the source text (FR), the machine translated output (MT), the post-edited output (PE) and the human translation (HT) (i.e. the reference), drawn from various WMT editions.
	The translations were produced by a \ac{PBMT} system (Moses). 
	The instructions to those performing the post-editing were to make the minimum amount of corrections necessary for a publishable translation \cite{Potet:2012}.
	We show results using the \ac{MT}-\ac{PE} documents, taking the PE version for comparison instead of the reference \ac{HT}. 
	Using the \ac{PE} for comparison will avoid mismatches that are due to variances (e.g. style) in freely created reference translations. \ac{HT} will of course include much greater stylistic variation, which our metric cannot easily capture, as will be explained later. Our hypothesis is that our metric should score the \ac{PE} more highly than the \ac{MT}, with many instances where both score equally, due to the fact that the \ac{MT} will correctly render the discourse connectives in those texts, 
	while other sentences in the source will not have explicit connectives at all.
	\paragraph{WMT Test Data}
	For comparison, we also score all the French-English submissions from the 2014 WMT shared translation task \cite{WMT:2014} with our model.\footnote{2014 being the last year of French-English submisions, as recent editions of WMT do not include French.} 
	We then show the correlations between the ranking of MT systems that participated in the shared task and our metric, in comparison with the best scoring reference-based metrics from the 2014 WMT metrics task \cite{bojar:2014:metrics}. The top metric in this shared task (DiscoTK \cite{Guzman}) also includes a discourse relation component, as described in Section \ref{related}.
	We take the French-English translations, comprising 175 documents, a total of 3,003 sentences per system submission, with eight system submissions in total. We also linearly combine our score with the top metrics, to establish whether the addition of our metric is complementary and results in an increase.
	
	\subsection{Discourse Connectives}\label{sec:connectives}

	\paragraph{Training}
	In French, many discourse cues are composed of several words \cite{roze_Lexconn}, such as \textit{au moment o\`{u}}. In fact, Laali \shortcite{Laali2014} found that in French larger order ngram connectives are more prevalent than in English and that, for example, LexConn contains 69 4-gram connectives.
	To a lesser extent this is true also in English (e.g. \textit{of course}, \textit{as well as}, \textit{in addition}). In order to have a phrase representing the full cue returned in our embeddings, we hyphenate the discourse cues in the training data. 
	To capture these discourse cues, we train bilingual word embeddings as per Luong \shortcite{luong2015bilingual}, using the MultiVec toolkit \cite{multivec} with the following modifications.
	We identify the French discourse connectives in our training corpus using Lexconn, and 
	then hyphenate all the identified discourse connectives before training bilingual embeddings. We did the same for the English corpus, hyphenating all the connectives which appeared in the list of 226 compiled by Knott \shortcite{Knott93}. 
	This ensures that for multi-word connectives, the full discourse cue is mapped.
	For example, for the French discourse connective \textit{parce\_que}, the model correctly returns \textit{because}, among others.
	We also train embeddings with a non-hyphenated version of the corpus, and with the full Europarl dataset \cite{Europarl}.
	We then back off to each of the above in turn (hyphenated, non-hyphenated, full Europarl), where the first model has no equivalent for our searched connective.
	The parallel data we use for our hyphenated embeddings consists of 214,972 sentence pairs, while the embeddings we trained with the full Europarl consists of 154,915,709 sentence pairs.
	
	\paragraph{Testing}
	Once we have identified the existence of a French discourse connective in the source text, 
	we then use syntactic cues based on findings by Laali \shortcite{Laali2014} to verify if the cue is being used in a discourse context.
	These involve identifying whether the potential connective in the source text has a syntactic tag of correct category (e.g. ADV, C, MWADV, MWC, CS etc.), for example that the word \emph{alors} is being used in a discourse sense, as part of \emph{alors que}, not simply as a comment word. 
	The same happens in English: the word \emph{and} can be used in a discourse sense (to join two clauses), or a non-discourse sense (as part of a listing).
	This helps us to determine if the lexical items from LexConn identified in the French source text are being used as a discourse connective. Laali \shortcite{Laali2014} determined that syntax could be used to filter out constituents that were not discourse connectives. This has previously been done successfully in English \cite{PitlerNenkova:2009}.

	\subsection{Discourse Relations}\label{sec:relations}
	Discourse relation theories include the hierarchical tree-based \ac{RST} \cite{mann1988} and lexically-based \ac{PDTB} \cite{Prasad:2008} 
	We base our experiment loosely on the latter.
	In \ac{PDTB} the intention is to identify a discourse relation between two arguments, without presupposing any hierarchical structure. Discourse relations can be implicit or explicit. If explicit, they are generally signalled by 
	discourse connectives. Implicit relations can be derived from the context, but have no given markers.
	We restrict this experiment to sentence-level relations, as did Guzm\'{a}n \shortcite{Guzman}, on the basis that \ac{MT} systems are and \ac{MT} evaluation are at sentence-level. 
	
	In the absence of a French discourse parser to identify the occurrence of a connective and give an indication of the discourse relation being used in the (French) source  text we use LexConn. 
	Any connectives occurring in our training data will be represented in the crosslingual discourse embeddings we create. 
	In addition to determining the discourse versus non-discourse usage of a particular word or phrase as described above, 
	we also use syntax to disambiguate connectives where there are multiple senses for the given connective. For example, in English the word \emph{since} can have either a \emph{causal} or a \emph{temporal} sense.  This has previously been done successfully in English \cite{PitlerNenkova:2009}. This applies in French too. 
	We identified 80 ambiguous connectives in LexConn which required additional disambiguation
	(out of a total list of 328 connectives usages), 
	for which we checked for occurrences in the ANNODIS corpus \cite{ANNODIS} of French discourse annotations 
	to ascertain their correct discourse relation in a given context. 
	Those potentially ambiguous connectives which did not occur we discarded, as we had no context to evaluate, moreover they were more likely to be infrequent. 
	We then filtered further, 
	where the two ambiguous relations came under the same higher level categorisation (such as for \emph{mais}, which can be indicative of the relations \emph{contrast} or \emph{violation}, and when mapped under our four broad mappings (see below) results in same category, \emph{comparison}).
	Our final list of ambiguous connectives requiring disambiguation included: \emph{apr\`{e}s}, \emph{aussi}, \emph{alors que}, \emph{depuis que}, \emph{en}, \emph{tandis que}, \emph{m\^{e}me}, \emph{si}, \emph{tout d'abord}.
	We then devised disambiguation rules based on heuristics or syntax for these remaining ambiguous connectives. For example, for the connective \emph{aussi}; if it was in sentence initial position and therefore upper case,  could be regarded as pertaining to \emph{result}, otherwise to discourse relation \emph{parallel}. 
	For most of the connectives captured in our embeddings there is no ambiguity, and we take the discourse relation as defined by LexConn. These rules are simply to cover the few ambiguous ones, to ensure that we compute our score against the correct discourse relation.
	
	For  assessing the discourse relation on the target side (English), we used the discourse tagger developed by Pitler \shortcite{PitlerNenkova:2009}, which identifies the top level of PDTB discourse relations, namely Temporal, Comparison, Contingency and Expansion. The relations of  LexConn (30 in total) can be manually mapped roughly to the second level of the PDTB, 
	so we manually establish the corresponding mappings, assigning relevant ones to the four PDTB ones which the tagger identifies. Ideally we would like to have a more fine-grained approach, but in the absence of a discourse parser for the French side, this was the most robust approach we could devise. Given the variation of discourse relations in LexConn, a more detailed mapping would be more difficult without a great deal of additional analysis. Moreover, discourse parsers still fail to reach high levels of accuracy for relationship identification, and so we opted for a more flexible approach in this initial experiment. 
	We make the assumption that if the tagger cannot identify a relation in the \ac{MT}, then it is probably not properly rendered, and so will not be scored. 
	
	\subsection{Dis-Score Metric}\label{sec:metric}
	Our metric, Dis-Score, is composed of the probability given to a potential discourse connective in English for any particular French connective from the specifically pretrained bilingual embeddings (DC), 
	combined with a score reflecting the correctness of the discourse relation match (DR), weighted by 
	$\gamma$. 
	We calculate the value for 
	$\gamma$ by doing grid search cross validation on the LIG corpus using the scikit-learn toolkit, \cite{scikit-learn}, which results 
	$\gamma$ to a value of 0.045. 
	This is summed over each sentence of the document, and normalised by the number of sentences in the document. Formally, the score is as follows, where Dis-Score(D) is our overall score for document D, $N$ is the number of sentences in the document, and $M$ is the number of discourse connectives in a sentence. $ST$ is the source text, and so the scoring function tracks the number of discourse connectives and relations in the French text.
	
	\begin{align}\label{scoring}
	\small
	\mbox{Dis-Score(D)}	& = \frac{1}{N} \sum_{i=1}^N \sum_{j=1}^M  \frac{DC}{DC_{ST}} * \gamma \frac{DR}{DR_{ST}}
	\end{align}

	\section{Results and Discussion} \label{DRresults}
	We evaluate our Dis-Score metric in several ways. Firstly, on the LIG dataset, we compute scores for MT and PE and hypothesise that the \ac{PE} should score higher than the \ac{MT}, although there will be many ties, where the \ac{MT} correctly renders the discourse relation, or where there is no explicit discourse relation in the source text. There may be instances where the \ac{PE} renders the discourse relation in an implicit manner, while this is unlikely to be the case for the \ac{MT}, and which therefore results in a situation where the \ac{MT} may sometimes score higher.
	Secondly, we evaluate the outputs from the WMT14 system submissions, and establish how our scores compare with the official system and segment rankings.
	We did try combining our scores with a \ac{QE} system, but the results were rather random, since 
	it depended on the exact partitioning of data into train and test. Moreover, the segment level score varies a great deal, depending on how many connectives are in that segment.
	
	
	\subsection{Results on LIG Test Set}
	
	In Table \ref{tab:results} we report our results on the LIG corpus comparing the metric scores for the \ac{PE} and the \ac{MT}. Under our metric, the scores for the \ac{PE} are greater or equal to the \ac{MT} for 281 of the 361 documents (78\%) in the LIG corpus. Half of the documents are tied according to the metric, where both documents score equally. This is to be expected, as the \ac{MT} successfully renders a significant amount of the connectives. Moreover, there may be sentences on the source side where there are no discourse connectives present:
for the LIG corpus as a whole, there are 2998 sentences where one or more  explicit discourse relation is detected, out of a total 10756 sentences. 	
	On closer analysis, for the documents where the \ac{MT} scored more highly than the \ac{PE}, sometimes the tagger failed to identify the connective in the \ac{PE}, despite recognising the same connective in the \ac{MT}. 	
	\begin{table}
		\centering
		\small
		\begin{tabular}{|l|r|r|}
			\hline
			Output 	& Number of Wins & Percentage of Wins \\
			\hline
			\textsc{PE}	&113/361&31\%\\
			\textsc{MT}	&80/361&22\%\\
			\hline		
		\end{tabular}
		\caption[Results for Dis-Score metric on LIG]{Number and proportion of times PE wins over the MT version in the LIG corpus according to the Dis-Score metric at the document-level. In 168 out of 361 cases, the metric score was the same for MT and PE.\label{tab:results}}
	\end{table}
	Interestingly, the \ac{MT} often scores better than the \ac{HT} under our model.
	We found that there are numerous instances where the relation is rendered in a more subtle manner in the \ac{HT}, and can be inferred from implicit discourse relations. For example, we found:
	\emph{mais cela n'aura servi \`{a} rien} translated as
	\emph{to no avail} in the \ac{HT}, which our model did not score. Here the \ac{MT} scored for having an explicit connective \emph{but}, which is the most probable connective for the French \emph{mais}.
	This supports the findings by Meyer \shortcite{WebberMeyer}, that up to 18\% of discourse connectives are not rendered explicitly in human translations. As such, they are missed by our current configuration, which does not take account of implicit relations.
	
	However, we consider our metric effective for measuring the extent to which the \ac{MT} output captures explicit discourse relations from the source text, and renders them in the target text. Due to the nature of \ac{MT}, it closely follows the source text. 
	This automatically increases the embeddings score, where \ac{MT} selects most probable equivalent, for cases where it has correctly translated it. Whereas even when the \ac{HT} does use an explicit discourse relation, such is the natural variability in human translation that the connective may not have been the most probable translation. For example, the English \emph{but} would have the highest probability score for French connective \emph{mais}, while \emph{however} or \emph{nevertheless} are equally good choices. 
	This is also the reason why we chose to compare the \ac{MT} with the \ac{PE}.

	\subsection{Results on WMT Test Set}
	For further evaluation, we follow Hajlaoui \shortcite{Hajlaoui:2013} and also show how the system submissions from 2014 WMT shared translation task score under our metric. 
	It should be noted however that we measure discourse relations in isolation, focussing solely on explicit connectives, whereas \ac{MT} output has other problems which will affect the WMT rankings. These more general problems are the target of most reference-based, n-gram matching metrics, such as BLEU \cite{Papineni:2002}. 
	Therefore, directly comparing our results to standard metrics is not entirely meaningful. On the other hand, by taking the overall ranking of the MT systems (generated from human evaluation) and checking how our metric and other metrics would rank the same systems, we can gain insights on where metrics fail or are complementary. Therefore, we also show two versions of DiscoTK, the top scoring metric in WMT14.

	\begin{table}
		\centering
		\small
		\begin{tabular}{|l|r|r|r|r|r|}
			
			\hline
			system 					&Dis-Score &Human ranking &DiscoTK-party& DiscoTK-light\\
			\hline
			
			\textsc{UEDIN}	 	 	&	0.437	(3)		&1			&	0.829	& 1	\\
			\textsc{STANFORD}		&	0.414	(7)		&2 			&0.768		&0.957\\
			\textsc{KIT}			&	0.414	(7)		&2 			&	0.756	&0.939\\
			\textsc{ONLINE-B}		&	0.417	(6)		&2 			&0.738		&0.855\\
			\textsc{ONLINE-A}		&	0.448	(2)		&3 			&0.651		&0.814\\
			\textsc{RBMT1} 			&	0.430	(4)		&4 			&	0.200	&0.227\\		
			\textsc{RBMT4} 			&	0.459	(1)		&5 			&	0.013	&0.047\\
			\textsc{ONLINE-C}		&	0.421	(5)		&6 			&	-0.063	&0.004\\
			
			\hline
			
		\end{tabular}
		\caption[Dis-Score and top WMT14 submissions]{Human ranking of 2014 WMT MT system submissions compared to Dis-Score and top WMT14 metric rankings. 
			\label{tab:disscore_wmtresults}}
	\end{table}
	
	Both DiscoTK-light and DiscoTK-party include discourse structure, which is not covered by other metrics. The latter was also the best performing at the WMT14 metrics task \cite{WMT:2014}. 
	DiscoTK-light combines variations of discourse structure from comparing \ac{RST} discourse trees of \ac{MT} and \ac{HT} using a convolution tree kernel, 
	while the DiscoTK-party metric combines the latter with other metrics operating at different levels (lexical, etc.) \cite{Guzman}. 
	As previously mentioned, DiscoTK parses the \ac{MT} output into \ac{RST} discourse trees and compares it to the \ac{HT} tree, whereas we compare the \ac{MT} with the \ac{ST}, and check whether the cues and semantics for a particular discourse relation are comparable (instead of comparing the structure). As such, they are complementary.
	As can be seen from the results in Table \ref{tab:disscore_wmtresults}, our metric would lead to a different ranking, where a rule-based system (RBMT4) would rank the highest. This is perhaps not surprising, given that rule-based MT systems tend to model linguistic structures (including discourse) more explicitly through rules. DiscoTK approximates the human ranking very well, but it should be noted that DiscoTK-party is combined with a number of other metrics, and uses machine learning models trained on human rankings. 
	\begin{table}
		\centering
		\small
		\begin{tabular}{|l|l|ll|}
			\hline
			\bf{Metric}   &  fr-en    \\
			\hline		
				Dis-Score & -0.213   \\
			\hline
			DiscoTK-light+DisScore  &0.969 (DiscoTK-light alone 0.965)\\
			\hline
			DiscoTK-party+DisScore & 0.975 (DiscoTK-party alone 0.970)\\
			
			\hline
		\end{tabular}
		\caption[Dis-Score correlation with human judgements]{Results on WMT14 at system level: Pearson correlation with human judgements. Our Dis-Score in linear combination with various DiscoTK 2014 WMT submissions. \label{tab:disscore_wmtcorrelations}}
	\end{table}
	\begin{table}
		\centering
		\small
		\begin{tabular}{|l|l|l|l|l|l|}
			\hline
			\bf{Metric}     &Average  &  wmt12 &   wmt13 &   xties \\
			\hline		
			Dis-Score     &0.012*     &-0.941* & 0.263* &  0.250*\\
			
			\hline
		\end{tabular}
		\caption[Dis-Score correlation with human judgements]{Results on WMT14 Fr-En at segment level: different variations of Kendall's $\tau$ rank correlation with human judgements.  \label{tab:disscore_wmtcorrelations_seg}}
	\end{table}
	We display the system-level correlations with human judgements, as per the Pearson correlation  for WMT14, and display results for our system in Table \ref{tab:disscore_wmtcorrelations}, as well as some of the others for comparison. Given that our metric only looks at one isolated discourse component, we do not expect the correlation to be high on its own. 
	We combine our scores linearly with DiscoTK variants as per Guzm\'{a}n \shortcite{Guzman}, discovering that when combined with DiscoTK-light, the resultant score is close to that of DiscoTK-party. In addition, when combined with DiscoTK-party, it ranks second metric overall for that language pair. It therefore is complementary as it increases the already high correlation, and could potentially be higher with tuning. 
	
	We also display the segment-level correlations with human judgements in Table \ref{tab:disscore_wmtcorrelations_seg}. 
	There are many segments with score of 0 since they have no detected connectives, and the per-segment variation of Dis-Score is high, since some sentences have several discourse relations, while others have none. Humans judgements consider other aspects and quality as a whole, whereas our metric just measures transfer of explicit discourse connectives. As such the correlation is low, at 0.012, but is positive. The highest segment-level correlation was 0.433. Under the Kendall's $\tau$ variant used for wmt13, which handles ties differently \cite{WMT:2014}, the correlation is higher at 0.263. The number of non-zero segments varies from system to system, but ranges from 504 to 566 segments out of a total of 3003.
	
	\begin{center}
		\begin{table}
			\small
			\begin{tabular}{|m{5em}|m{13cm}|}
				\hline
				MT system& Translation \\
				\hline
				source&``Tout est mis en oeuvre pour que le Charles-de-Gaulle puisse faire son d\'{e}ploiement pr\'{e}vu en fin d'ann\'{e}e 2013", explique cependant DCNS.\\
				\hline
				RBMT1& ``All is implemented {\em so that} Charles-of-Gaulle can make his deployment envisaged at the end of the year 2013," explains DCNS however.\\
				\hline
				STANFORD& ``Everything is being done to the charles-de-gaulle could make his planned deployment at the end of the year 2013,'" explains however DCNS.\\
				\hline
				UEDIN& ``Everything is being done to the charles-de-gaulle could make his planned deployment at the end of the year 2013," explains however DCNS.\\
				\hline
				ref&``Everything has been put in place to enable the Charles-de-Gaulle to be deployed as planned at the end of 2013," explains DCNS.\\
				\hline
			\end{tabular}
			\caption[Examples of MT system outputs]{\label{tab:examples} Examples of translations from different MT systems, with the marker {\em so that} correctly preserved by RBMT1.
			}
		\end{table}
	\end{center}
	Comparing the rankings (Table \ref{tab:disscore_wmtresults}), the main difference is that the rule-based systems do better under our model, and move from the lower half of the table to the top. 
	To give an intuition of why this is the case, we include some examples in Table \ref{tab:examples}. For this sentence, the RBMT1 system scored for having \emph{so that} as a translation of the French \emph{pour que}, which was recognised as discourse relation \emph{goal} which correctly mapped to Contingency. The other two systems (which are among the highest ranking) do not have this discourse cue, which is important for the understanding of the text. While the construct \emph{to + infinitive} would be equivalent, neither the STANFORD or UEDIN have an infinitive following the \emph{to}.
	As can be seen from the reference, the discourse relation is implicit in the human translation, and rendered by \emph{to enable}.
	
	In fact the human judgements on WMT are not themselves at document level, and so are
	not therefore directly comparable. Humans evaluations at WMT are on a window of a couple of source sentences, with no target translation context, and therefore do not give credit to models which overall may have a more consistent or coherent output at document level.
	\subsection{Discussion}
	Hoek \shortcite{hoek-eversvermeul-sanders:2015:DiscoMT} found that both implicitation and explicitation of discourse relations occurs frequently in human translations, i.e. explicit relations made implicit and vice versa. 
	In addition, more research needs to be done to ascertain how often discourse connectives in French are not directly rendered in English. 
	Ideally, a well-tuned model should take account of the amount of implicitation and explicitation typical of the language pair in question. 
	Finally, using LexConn to identify the discourse connectives in the source text is occasionally problematic, as these cues are open class words and, as already noted \cite{Laali2014}, not all in LexConn. 
	
	
	\section{Conclusions} \label{conclusions}
	We have shown how the crosslingual transfer of discourse relations can be measured in \ac{MT}, in terms of connectives as cues signalling the discourse relations, as well as assessing the subsequent semantic transfer of the discourse relation. 
	Our model is measured against the source text and does not use a gold reference or alignments. 
	For more flexible and realistic evaluation of a translation, there is a growing need to move away from the current approach towards assessing the translated output conditioned on the source text. 
	This will need to be a multifaceted semantic approach, of which assessing the transfer of discourse relations from source to target is but one element which requires evaluation. 
	
	Our work introduces a way in which this can be done, successfully scoring the \ac{PE} greater or equal to the \ac{MT} 78\% of the time.
	We believe it is novel, in that we do this using crosslingual word embeddings pretrained for multiword discourse connectives, and incorporate discourse relation mappings between source and target text. 
	It does reward translations that are closer to the source than some better human translations, but this makes it suited to evaluating \ac{MT} which by its very nature is more similar to the source text than human translations.
	By necessity, it is dependent on a parser and tagger, which do not always correctly assess the constituents or discourse relations. Ultimately using a discourse parser would be better, as it would reduce the bias from the higher probability assigned to frequent connectives,
	and particularly as we could then map the discourse relations at a more detailed level (beyond first level PDTB). 
	While we recognise that this initial experiment only covers explicit, not implicit relations, and intrasentential not intersentential ones, 
	we believe it is a novel and constructive effort to address an evaluation gap. 
	

	\bibliographystyle{acl}

	\bibliography{Karin}
\end{document}